\documentclass{article}
\usepackage[T1]{fontenc}

\usepackage{microtype}
\usepackage{graphicx}
\usepackage{subcaption}
\usepackage{booktabs} 
\usepackage{hyperref}

\usepackage[preprint]{icml2026}

\usepackage{amsmath}
\usepackage{amssymb}
\usepackage{mathtools}
\usepackage{amsthm}

\usepackage{algorithm}
\usepackage{algorithmic}

\usepackage{tabularx}
\usepackage{multirow}
\usepackage{amsfonts}
\usepackage{nicefrac}
\usepackage{tikz}
\usepackage{xspace}
\usepackage{xcolor}
\usepackage{listings}

\usepackage[most]{tcolorbox}
\usepackage{enumitem}

\usepackage[capitalize,noabbrev]{cleveref}

\lstdefinestyle{simprompt}{
  basicstyle=\ttfamily\scriptsize,
  breaklines=true,
  breakatwhitespace=true,
  breakindent=0pt,
  columns=fullflexible,
  keepspaces=true,
  showstringspaces=false,
  aboveskip=0pt,
  belowskip=0pt,
}

\newtcblisting{promptcard}[1]{
  enhanced,
  breakable,
  listing only,
  listing options={style=simprompt},
  colback=black!4,
  colframe=black!22,
  colbacktitle=black!10,
  coltitle=black!80,
  fonttitle=\sffamily\footnotesize\bfseries,
  boxrule=0.35pt,
  leftrule=1.6pt,
  arc=1.5pt,
  left=5pt,
  right=5pt,
  top=5pt,
  bottom=5pt,
  toptitle=2pt,
  bottomtitle=2pt,
  title={#1},
}

\newtcolorbox{promptnotecard}[1]{
  enhanced,
  breakable,
  colback=black!4,
  colframe=black!22,
  colbacktitle=black!10,
  coltitle=black!80,
  fonttitle=\sffamily\footnotesize\bfseries,
  fontupper=\ttfamily\scriptsize,
  boxrule=0.35pt,
  leftrule=1.6pt,
  arc=1.5pt,
  left=5pt,
  right=5pt,
  top=5pt,
  bottom=5pt,
  toptitle=2pt,
  bottomtitle=2pt,
  title={#1},
}

\usetikzlibrary{positioning,arrows.meta,shapes,calc,fit,decorations.pathreplacing}
\usetikzlibrary{arrows.meta, positioning, fit, backgrounds, shapes.geometric, calc}

\definecolor{prodbd}{HTML}{5C4E8A}
\definecolor{prodbg}{HTML}{E6E4ED}
\definecolor{simbd}{HTML}{E07A3D}
\definecolor{simbg}{HTML}{FAEBE2}
\definecolor{intentbd}{HTML}{3D6B99}
\definecolor{intent}{HTML}{E2E9F0}
\definecolor{accent}{HTML}{5E9E8E}
\definecolor{hl}{HTML}{E7F0EE}
\definecolor{planbd}{HTML}{5C4E8A}
\definecolor{plan}{HTML}{E6E4ED}
\definecolor{ink}{HTML}{1B2430}
\definecolor{edge}{HTML}{6E6A78}
\definecolor{clbg}{HTML}{F0F1F3}
\definecolor{clbd}{HTML}{B8BDC4}

\tikzset{
  box/.style   = {rectangle, rounded corners=2pt, draw=accent, line width=.5pt,
                  fill=white, align=center, font=\scriptsize, inner sep=4.5pt,
                  text=ink, minimum height=8mm},
  hlbox/.style = {box, draw=accent, fill=hl},
  intbox/.style= {box, draw=intentbd, fill=intent},
  planbox/.style={box, draw=planbd, fill=plan},
  simbox/.style ={box, draw=simbd, fill=simbg},
  dec/.style   = {diamond, aspect=2.1, draw=accent, fill=white, align=center,
                  font=\scriptsize, text=ink, inner sep=1pt},
  arr/.style   = {-{Stealth[length=2mm,width=1.6mm]}, draw=edge, line width=.7pt},
  darr/.style  = {arr, dashed},
  lbl/.style   = {font=\scriptsize\itshape, text=edge, inner sep=1.5pt},
  cl/.style    = {rounded corners=4pt, draw=clbd, fill=clbg, inner sep=6pt},
  cltitle/.style={font=\scriptsize\bfseries, text=accent},
}

\graphicspath{{images/}}

\theoremstyle{plain}

\theoremstyle{definition}

\theoremstyle{remark}

\definecolor{SeaGreen}{RGB}{46, 139, 87} 

\newcommand{\snowglobe}{\textsc{Snowglobe}\xspace}
\DeclareRobustCommand{\cdv}[1]{\ensuremath{V#1_{\mathrm{CD}}}}
\DeclareRobustCommand{\cmv}[1]{\ensuremath{V#1_{\mathrm{CM}}}}

\icmltitlerunning{Simulation for Production Customer Experience AI Agents at 140M Scale}

\begin{document}
\raggedbottom

\twocolumn[
  \icmltitle{Screen Before You Serve:\\Simulation for Production Customer Experience AI Agents at 140M Scale}

  \icmlsetsymbol{equal}{*}

    \begin{icmlauthorlist}
      \icmlauthor{Edesio Alcobaça}{equal,nubank}
      \icmlauthor{Kevin Rossell}{equal,nubank}
      \icmlauthor{Aman Gupta}{nubank}
      \icmlauthor{Shao Tang}{nubank}
      \icmlauthor{Jiwoo Hong}{nubank}
      \icmlauthor{Pabel Carrillo-Mendoza}{nubank}
      \icmlauthor{Wanderson Conceição Ferreira}{nubank} \icmlauthor{Alvaro Tedeschi}{nubank}
      \icmlauthor{Zayd Simjee}{guardrails}
      \icmlauthor{Shreya Rajpal}{guardrails}
      \icmlauthor{Bruno Finardi Hime}{nubank}
      \icmlauthor{Christian Sousa}{nubank}
      \icmlauthor{Luis Moneda}{nubank}
      \icmlauthor{Herbert Fei}{nubank}
      \icmlauthor{Daniel Silva}{nubank}
      \icmlauthor{Rohan Ramanath}{nubank}
    \end{icmlauthorlist}
    
    \icmlaffiliation{nubank}{Nubank}
    \icmlaffiliation{guardrails}{Guardrails AI}
    
  \icmlcorrespondingauthor{Aman Gupta}{\texttt{aman.gupta@nubank.com.br}}
  \icmlcorrespondingauthor{Shao Tang}{\texttt{tang.shao@nubank.com.br}}

  \icmlkeywords{Customer experience agents, large language models, user simulation, agent evaluation, open-weight models}

  \vskip 0.3in
]

\printAffiliationsAndNotice{\textsuperscript{*}Joint first authors. }  

\begin{abstract}
Customer experience (CX) agents use tools and large language models to address customer requests and guide conversational interactions with an organization's products. Improving these agents, especially in regulated industries, is difficult: they must detect intent, follow complex operational policies and use tools reliably. Manual end-to-end testing offers limited coverage, while live experiments expose customers to failures that can erode trust.

We present a hypothesis-driven simulation workflow for screening candidate CX agents before deployment. Synthetic customers react to agent responses and simulated tool outputs enable multi-step agentic workflows without invoking production backends. We use the Snowglobe simulator on Nubank's Card Delivery agent and its expanded successor, Card Management - Nubank’s highest-volume chat-support agent in Brazil. Across \textbf{4} deployed versions, simulated and production version-level binary evaluator scores show high correlation.  Simulation-guided iteration increased transactional net
promoter score (tNPS) by \textbf{36.69 points} in a live A/B test. We also screened open-weight configurations in over \textbf{16,000} simulated conversations. In a subsequent live A/B test, the selected model increased self-service rate (SSR) by \textbf{8.82} percentage points to the highest level observed at Nubank, with no statistically significant change in  tNPS. Simulation made broad exploration of models, reasoning settings, and prompts feasible without customer exposure, enabling production improvements that would have been impractical to pursue through live experimentation alone.
\end{abstract}

\begin{figure*}[!htb]
  \centering
  \includegraphics[width=1.0\linewidth]{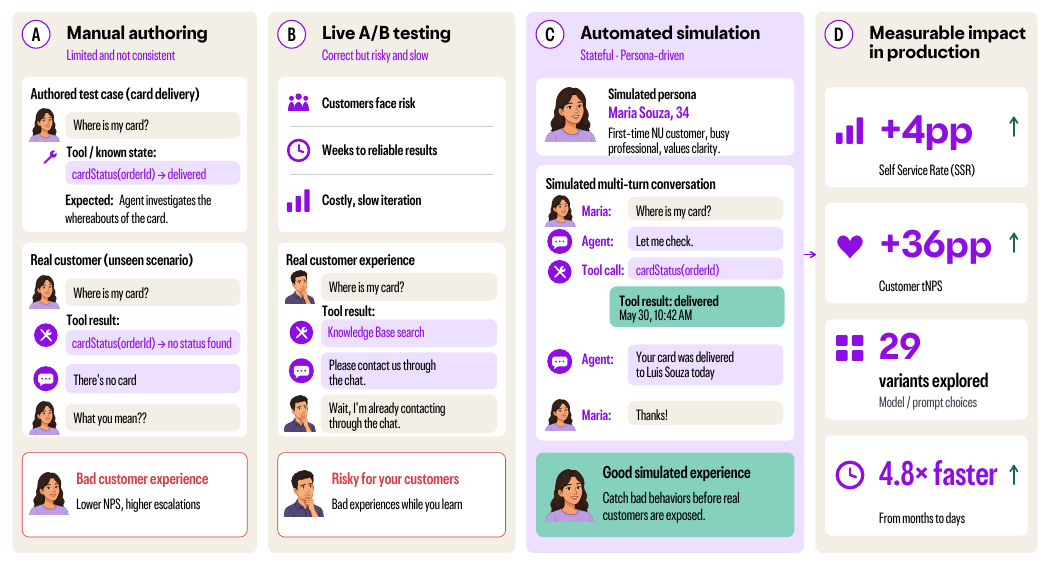}
  \caption{\textbf{Three ways to decide whether an agent revision is safe to
  ship.} Manual authoring (A) exercises the scenarios someone thought to write
  down; some tool states that real customers encounter may be missing. A live A/B test (B) measures outcomes directly in production but exposes customers to the agent changes being tested. On-policy,
  tool-boundary simulation (C) answers the tool call synthetically and lets the
  agent act on its own policy, so multi-turn failures can surface before customers are exposed. In production (D) the screened agent moved self-service rate and
  transactional NPS against a matched comparison.}
  \label{fig:overview}
\end{figure*}

\section{Introduction}
\label{sec:intro}

\begin{figure*}[ht]
  \centering
  \includegraphics[width=1.0\linewidth]{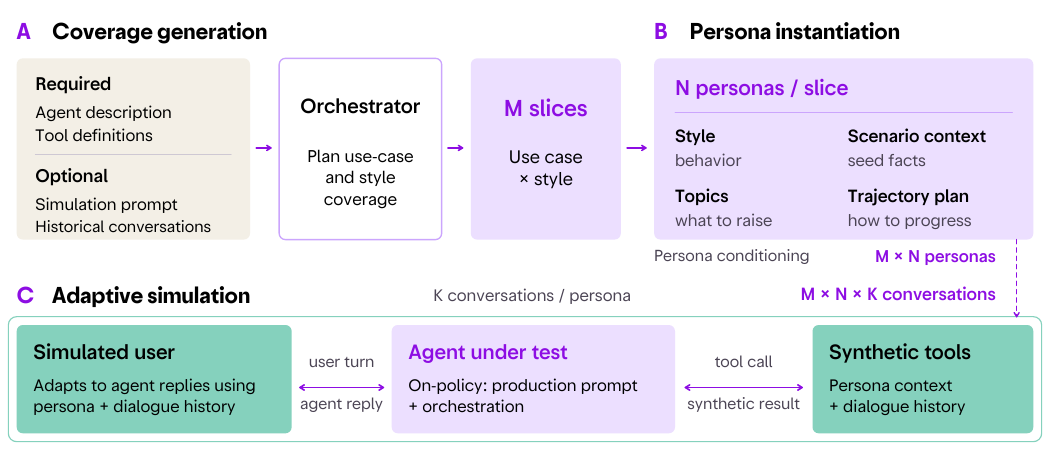}
  \caption{End-to-end trajectory generation with
  \snowglobe: an orchestrator reads the inputs and plans coverage over
  use-case and style slices; each slice generates personas carrying
  style, state, topics, and trajectory; each persona holds \textbf{one} or more
  multi-turn conversations with the agent under test, whose tool calls
   are answered by a mock tool layer.}

  \label{fig:snowglobe}
\end{figure*}

Recent advancements in large language models (LLMs), including complex reasoning \citep{guo2025deepseek,openai2025o3o4mini}, multi-step tool use \citep{qin2024toolllm,schick2023toolformer}, and long-context understanding \citep{geminiteam2024gemini15}, have improved the real-world applicability of LLMs. By perceiving their environment and using tools to address users' queries, LLM-based agents are being deployed for both personal and business applications \citep{luo2025amazon,gupta2026building,openclaw2026,research2026composer2technicalreport}. While AI-assisted workflows have become easier to adopt in the era of agents, building reliable benchmarks for domain-specific, open-ended tasks remains challenging \citep{lynch2025agentic,zhu2026establishing}.

This challenge is especially pronounced for customer experience (CX) agents, which are among the most widely adopted agentic applications across businesses, with deployments at companies such as Airbnb \citep{zhao2025agent}, Amazon \citep{luo2025amazon}, and Nubank \citep{gupta2026building}. We define CX agents to include both \textit{reactive agents} that address customer questions, complaints, and support needs, and \textit{proactive agents} that anticipate customer needs and guide customers through conversational workflows to accomplish tasks using a company’s products and services. Although CX agents may not require the same depth of reasoning as coding or mathematical agents, they must meet complex, multidimensional requirements across diverse user groups. More specifically, CX agents should support personalization and localization \citep{kirk2024the}, respond appropriately to user intent across multi-turn conversations \citep{tack2026llmslostevolvinguser}, and handle sensitive data \citep{vijayvargiya2026openagentsafety,tien2026roguemisalignedagentbehavior}, among other requirements. Thus, the bars for safety and qualitative performance are higher than those for general agentic applications. Such unique expectations for CX agents raise a central question: \textbf{\emph{how can we establish that an agent is operational, handles edge cases, and is ready for safe online deployment?}}

Figure~\ref{fig:overview} contrasts \textbf{three} approaches:
(1) authoring test cases by hand, (2) exposing real customers in a live A/B test, and (3)
simulating customers on-policy at the tool boundary.
Although the first \textbf{two} options incorporate human feedback, they are slow and can provide inconsistent signals with small sample sizes. User simulation with synthetic personas has therefore been studied as an alternative for approximating the human preference distributions with LLMs \citep{ge2025scaling,naous2026flipping,li2026matraixsimulatingworld83}. However, limited controllability and the risk of score inflation raise concerns about discrepancies between simulated and real-world scenarios \citep{zhu2024howreliable,mehri2025ugst}.

In this paper, we explore \emph{user simulation} as an intermediate evaluation layer in which candidate agents are exercised through synthetic customer interactions before live deployment. Through case studies of the Card Delivery and Card Management agents at Nubank, we propose a \emph{hypothesis-driven, tool-boundary simulation recipe} for screening candidate agents and demonstrate its real-world business impact. Simulation has become an essential part of our CX agent development lifecycle, enabling broad exploration of models, prompts, and reasoning settings that would be impractical to pursue through live experimentation alone. Our contributions are as follows:
\begin{enumerate}
    \item \textbf{User simulation as screening-before-deployment}: We propose a simulation-aided agent deployment recipe with a hierarchical workflow connecting offline evaluation to online deployment.
    \item \textbf{Similarity analysis against real-world conversations}: We demonstrate moderate conversation-level similarity between real-world and simulated conversations, with the lowest cosine distance of \textbf{0.035}. Across four deployed versions, simulated and production version-level evaluator scores show high correlation.
    \item \textbf{Efficient development-to-production lifecycle}: The user simulation layer allows \textbf{4.8} times faster iteration of the development, screening, and deployment cycle compared with a workflow without a user simulation layer.
    \item \textbf{Production impact across agent and model changes}:
    The Card Management agent developed through simulation-guided
    iteration improved self-service rate (SSR) by
    \textbf{4.9 percentage points} and transactional net promoter
    score (tNPS) by \textbf{36.69 points} relative to Card Delivery
    in a live A/B test.
    \item \textbf{Generalizability across models}: We use the proposed simulation recipe to select Qwen3.5-122B-A10B as a replacement for the incumbent Card Management model. In a subsequent live A/B test, the replacement improved SSR by \textbf{8.82 percentage points} and reduced p95 latency by \textbf{25\%}, with no statistically significant change in tNPS.

\end{enumerate}

\section{Related Work}
\label{sec:related}

\paragraph{Customer experience agents in production.} An increasing number of companies across different business categories are adopting agentic workflows for customer support, including Amazon \citep{luo2025amazon}, Airbnb \citep{zhao2025agent,su2025llm}, Kakao \citep{park2025practical}, Alibaba Group \citep{jiang2025chatmap}, Thomson Reuters \citep{jucla2026retrieval}, and Nubank \citep{gupta2026building}. \citet{zhao2025agent} and \citet{su2025llm} propose human-in-the-loop feedback collection strategies and synthetic data generation pipelines for sustainably aligning customer experience agents with human preferences at Airbnb. \citet{gupta2026building} present a hierarchical pipeline spanning offline validation and online deployment for customer experience agents at Nubank, improving agent self-service rate (SSR) and transactional net promoter score (tNPS) \citep{reichheld2003nps}.

\begin{table*}[!t]
  \centering
  \caption{Card Delivery (CD) versions used to characterize the
  simulator and Card Management (CM) configurations evaluated
  through the simulation workflow.}
  \label{tab:agent-versions}
  \small
  \renewcommand{\arraystretch}{1.12}

  \begin{tabularx}{\textwidth}
    {@{}l>{\raggedright\arraybackslash}X@{\hspace{1.5em}}l>{\raggedright\arraybackslash}X@{}}
    \toprule
    \multicolumn{2}{@{}l}{\textbf{Card Delivery (CD)}} &
    \multicolumn{2}{@{}l@{}}{\textbf{Card Management (CM)}} \\
    \cmidrule(r){1-2}\cmidrule(l){3-4}
    \textbf{Version} & \textbf{Configuration} &
    \textbf{Version} & \textbf{Configuration} \\
    \midrule
    \cdv{1} & GPT-4.1; tools for customer-specific logistics queries. &
    \cmv{1} & GPT-5.2; CD prompts and tools with ReAct-style tool
    definitions~\citep{yao2023react}; baseline. \\
    \addlinespace[2pt]
    \cdv{2} & GPT-4.1; card selection and prevention of recent duplicate reissues. &
    \cmv{2} & GPT-5.2; Manual revision of \cmv{1}'s prompt and tool-use policy. \\
    \addlinespace[2pt]
    \cdv{3} & GPT-5.1; prompt rules for routing, personalization, and concise responses. &
    \cmv{3} & GPT-5.2; \cmv{1}'s prompt and tools with a different architecture. \\
    \addlinespace[2pt]
    \cdv{4} & GPT-5.2; tools to retrieve and update the registered delivery address. &
    \cmv{4} & GPT-5.2; Manual revision of \cmv{3}'s prompt and tool-use policy. \\
    \addlinespace[2pt]
    & & \cmv{5} & GPT-5.2; Return to ReAct; prompt revised using simulated examples. \\
    \bottomrule
  \end{tabularx}
  
\end{table*}

\paragraph{User simulation in benchmarking agents.} To benchmark agents across a wide range of domains, synthetic personas have been studied as a method for simulating diverse human demands, either at scale through verbalized prompts \citep{ge2025scaling,li2026matraixsimulatingworld83} or by fine-tuning the language model used for simulation \citep{naous2026flipping}. Recent agent benchmarks, in particular, leverage prompt-guided user personas to simulate real-world scenarios \citep{yao2024taubench,tan2025personabench,qian2025userbench,barres2025tau2bench,huang2026crmarenapro}. Specifically, $\tau$-Bench defines multi-turn scenarios in which agents interact with prompt-driven user simulators in retail and airline customer experience (CX) domains \citep{yao2024taubench}, while $\tau^2$-Bench extends this setting so that both agents and users can invoke tools, enabling more human-like actions by simulated users \citep{barres2025tau2bench}. Building on increasing efforts to expose CX agents to more realistic environments, we address the complementary question of how to characterize and use a simulator as a restricted screening layer in agent development for real-world deployment.

\paragraph{Reliability in user simulation.} Despite the increasing use of user simulation in agent benchmarking, a simulator of unknown fidelity can actively mislead \citep{zhu2024howreliable,mehri2025ugst}. Prompt-driven user simulators are often prone to score inflation through data leakage \citep{zhu2024howreliable} or may fail to adhere to their assigned goals in multi-turn scenarios \citep{mehri2025ugst}. Accordingly, rigorously validating the robustness of user simulation pipelines and their alignment with real human behavior remains challenging \citep{dou2025simulatorarena,zhu2026realusersim}. SimulatorArena explicitly measures agreement between human judgments and assistant ratings on public tasks such as math tutoring \citep{dou2025simulatorarena}, while RealUserSim grounds simulator personas in behavioral profiles extracted from real conversations and tests fidelity using an LLM-judged paired-trajectory Turing test \citep{zhu2026realusersim}. We characterize the reliability and robustness of a user simulation recipe by comparing simulated and real-world conversations, and demonstrate how simulation-based findings can inform changes that are subsequently evaluated through real-world business outcomes at Nubank. \textbf{Most importantly, our results demonstrate that imperfect simulation can still guide directionally correct improvements to production agents, with benefits confirmed with live A/B  evaluation.}

\section{Simulation Setup}

\label{sec:simulation}

This section consolidates the experimental setup for multi-turn agent trajectory generation and evaluation (Figure~\ref{fig:snowglobe}). Our pipeline selects a target agentic task, simulates on-policy conversations with synthetic personas and tool responses, and evaluates the resulting trajectories offline.

\paragraph{Target agents.}
We study Nubank's Card Delivery (CD) agent and its expanded successor, Card Management (CM), which replaced CD and supports a broader range of card-lifecycle requests. We characterize the simulation pipeline using CD traces and apply the resulting workflow to CM.

\subsection{Simulation pipelines}

\paragraph{Proposed simulation pipeline.} Our proposed trajectory-generation pipeline uses \snowglobe as its primary simulator \citep{guardrails2025snowglobe}. \snowglobe instantiates the persona-driven simulation pattern of \citet{park2023generative} and takes an agent description and tool definitions, optionally augmented with a simulation prompt and historical conversations. From these inputs, it constructs use cases, user profiles, tool relationships, and trajectory plans (Appendix~\ref{app:inference}). An orchestration agent allocates conversations across use-case and interaction-style conditions, after which the generated personas interact with the target agent over multiple turns. Each synthetic user turn is conditioned on the agent's preceding response, while tool calls are intercepted and answered with scenario-conditioned synthetic results rather than invoking production backends. A conversation ends when the persona's objective is achieved, judged unreachable, or subject to a turn or tool-specification limit. We adopt \snowglobe for its structured orchestration, controllable scenario coverage, and cross-call tool-state consistency support for on-policy, hypothesis-driven agent screening.

\paragraph{Naive LLM baseline.} To assess what can be achieved through direct prompting without \snowglobe's structured orchestration, we implement a naive LLM-based simulator. A persona language model generates informal Brazilian Portuguese customer turns, while a second language model produces synthetic results for the agent's tool calls. Both roles use \texttt{gpt-5.6-sol} with high reasoning effort and receive the same categories of non-test-set context as \snowglobe: the agent description, tool schemas, product-requirements meta-knowledge, and tool-call examples. Both simulators also use the same seeded scenario distribution. A baseline conversation ends when the persona signals completion, the agent transfers the conversation to a human, or the customer has sent \textbf{ten} messages. Unlike \snowglobe, the baseline does not orchestrate use-case and interaction-style coverage, generate structured trajectory plans, or use an agent-profile model to maintain cross-call consistency. Appendix~\ref{app:baseline-prompts} provides the baseline prompts.

\subsection{Card Delivery Agent}
\label{subsec:card_delivery}

\paragraph{Agent versions and data.} \textbf{8,000} real-world conversations between the users and the CD agent were evenly split into \textbf{two} sets: a development set to configure the simulators (\emph{i.e.,} simulation characterization) and a test set to validate the agent with the characterized simulation profiles. The \textbf{four} deployed CD versions are summarized in Table~\ref{tab:agent-versions} (left).

\begin{figure*}[!t]
  \centering
  \includegraphics[width=1.0\linewidth]{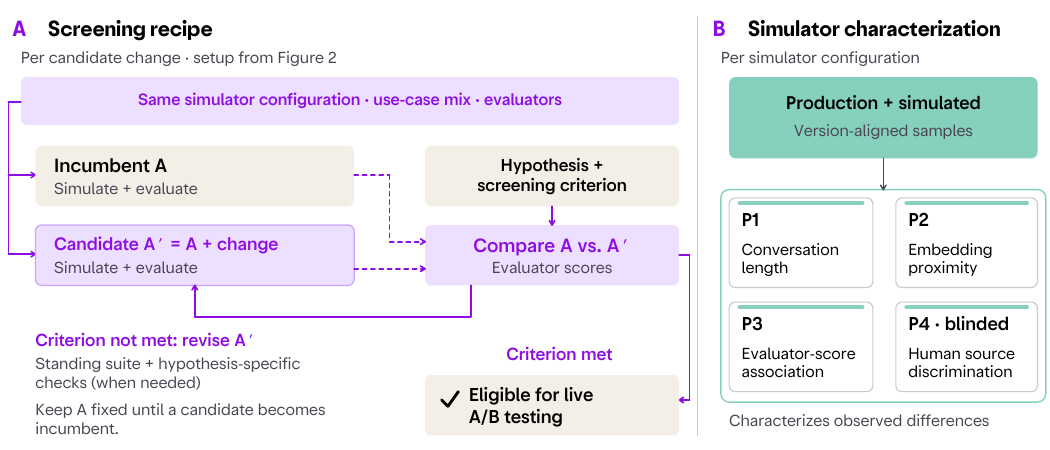}
  \caption{Simulation recipe and characterization: (a) candidate
  agent changes are simulated and compared with an incumbent baseline
  using standing and hypothesis-specific criteria to screen
  candidates, (b) \textbf{four} diagnostics compare variant-aligned production
  and simulated Card Delivery samples, characterizing observed
  differences without establishing backend fidelity.}
  \label{fig:recipe}
\end{figure*}

The test set contains \textbf{1,000} conversations from each agent version. We generated \textbf{250} synthetic trajectories via the simulator for each version.

\paragraph{Offline evaluation.} We use \textbf{five} semantic categories to evaluate agent trajectories offline. They are evaluated via LLM-as-a-Judge, using GPT-4.1-Mini\footnote{\url{https://openai.com/index/gpt-4-1/}}: (E1) Card reissue failure; (E2) Customer input verification; (E3) Card delivery data check; (E4) Response conciseness; (E5) Resolution conciseness following \citet{gupta2026building}. We evaluate each category with separate system prompts optimized via GEPA \citep{agrawal2025gepa}, which return binary pass/fail scores. 

\subsection{Card Management Agent}

\paragraph{Agent versions and data.}
We apply the simulation workflow shown in
Figure~\ref{fig:recipe}(a) to guide iterative improvements to
the CM agent, which uses GPT-5.2 as its backbone model.
Table~\ref{tab:agent-versions} (right) summarizes the baseline
and \textbf{four} additional versions developed through the proposed
simulation workflow.

\paragraph{Offline evaluation.} We used a single binary judgment from LLM-as-a-Judge to test the explicit hypothesis in each version as described above. Given the CM agent's execution traces and available tools, LLM-as-a-Judge provides binary feedback on whether a transfer to human support was unnecessary. We compare each candidate’s failure rate with that of \cmv{1} to decide whether to accept the change.

\paragraph{Online evaluation.} We assess production performance using \textbf{two} online metrics: transactional net promoter score (tNPS) and self-service rate (SSR). tNPS is the percentage of promoters minus the percentage of detractors in a post-interaction recommendation survey \citep{reichheld2003nps}.  tNPS is thus a measure of customer satisfaction and happiness. Self-service rate (SSR) is the proportion of sessions in which users completed the case with the agent without asking for human support.

\begin{figure*}[!t]
  \centering
  \includegraphics[width=1.0\linewidth]{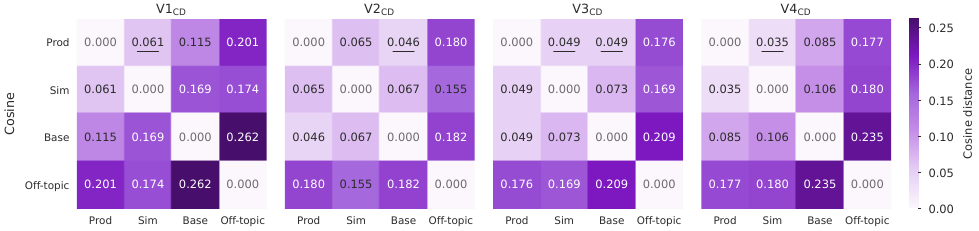}
  \caption{P2 -- \textbf{In all versions, both simulated groups are
closer to production than the off-topic control; \snowglobe is no
farther from production than the baseline in three versions.} Cosine
distances between whole-conversation embedding centroids. Underlined values
mark the nearest non-production centroid.}
  \label{fig:p2-centroid}
\end{figure*}

\section{User Simulation Recipe for CX Agents}
\label{sec:recipe}

Algorithm~\ref{alg:screening} uses tool-boundary trajectory
simulation for \emph{hypothesis-driven screening}~\cite{balog2024usersimulation}.
Its purpose is to expose beneficial or harmful candidate--incumbent
differences large enough to affect a deployment decision. Therefore, exact replication of production metrics is unnecessary. 

\begin{algorithm}[t]
\caption{Hypothesis-driven candidate screening}
\label{alg:screening}
\small
\begin{algorithmic}[1]
\REQUIRE Incumbent $A$; simulator configuration $S$;
         evaluation suite $E$; prespecified screening criterion $C$
\STATE $T_A \leftarrow \mathrm{Simulate}(A; S)$
\FOR{each candidate revision $\Delta$}
  \STATE $A' \leftarrow A + \Delta$
  \STATE $T_{A'} \leftarrow \mathrm{Simulate}(A'; S)$
  \STATE Evaluate every trajectory in $T_A$ and $T_{A'}$ with $E$
  \IF{the candidate--incumbent comparison meets $C$}
    \STATE Mark $A'$ eligible for live A/B testing
  \ELSE
    \STATE Reassess or revise $\Delta$
  \ENDIF
\ENDFOR
\end{algorithmic}%

\end{algorithm}

The algorithm first generates on-policy incumbent trajectories
using simulator configuration $S$, which specifies the target use-case
mix. These trajectories remain the shared baseline throughout the
screening round. Configuration uses the agent description and tool
definitions, optionally supplemented by a simulation prompt and
historical data.
Record scenario composition and keep distribution runs separate from
targeted probes. Scenario-conditioned tool outputs must match
declared schemas and remain consistent across calls within a
conversation. For example, a record queried twice should not change
state without an intervening action. Simulation exercises
dialogue--tool interactions without testing backend implementations
(Appendix~\ref{app:mocking}).

Each screening round is guided by a falsifiable hypothesis
and a prespecified screening criterion $C$. The hypothesis determines the behavioral
effect of interest, while $C$ defines how evaluator results
support eligibility for live testing. The evaluation suite $E$ comprises evaluators relevant to the
task and the hypothesis, including LLM-as-a-judge evaluators when
appropriate. It covers selected standing criteria and the proposed
behavioral effect. The same suite is applied to incumbent and
candidate trajectories; any newly introduced evaluator must also
score the baseline trajectories.
The standing CD suite was calibrated against human
annotations~\cite{gupta2026building}; calibration of each new
hypothesis-specific judge must be reported separately.

Candidate simulations hold $S$ fixed. Candidates meeting $C$
become eligible for live A/B testing; others are reassessed or
revised against the same incumbent.
The baseline is updated only after a candidate becomes the incumbent.
For candidates meeting $C$, smaller or ambiguous differences are left for live experiments
to resolve. This loop complements item-level evaluation and live A/B
tests; Figure~\ref{fig:recipe} summarizes it alongside the separate
simulator-characterization diagnostics.

\subsection{Simulator characterization}
\label{sec:framework}

We introduce \textbf{four} different diagnostics for the characterized simulators, including \textbf{three} automated evaluations and \textbf{one} human annotation session, namely P1 to P4. For \textbf{three} synthetic evaluations, we use the data and offline evaluation pipeline introduced in Section \ref{subsec:card_delivery}, \emph{i.e.,} each uses \textbf{1,000} held-out production conversations with real-world users and \textbf{250} \snowglobe traces per CD version. The diagnostics address complementary properties (Figure~\ref{fig:recipe}b), none as a standalone criterion.

\paragraph{P1: Conversation-length statistics.}
We compare the number of words in user messages and the number of user
turns per conversation. We report cumulative threshold percentages with full distributions in Appendix \ref{app:p1-dists}.

\paragraph{P2: Embedding diagnostics.}
We embed each conversation using \texttt{text-embedding-3-large}\footnote{\url{https://developers.openai.com/api/docs/models/text-embedding-3-large} with \texttt{user:} and \texttt{agent:}
prefixes}. Then, we measure cosine and Euclidean distances between the group centroids, and plot the \textbf{two}-dimensional Uniform Manifold Approximation and Projection \citep[UMAP]{mcinnes2018umap} visualization.

\paragraph{P3: Evaluator-score association.}
We apply \textbf{five} canonical evaluation categories, E1 to E5, in Section \ref{subsec:card_delivery} to the production, \snowglobe, and baseline pools. For each agent version, we report the average scores per category and their \textbf{95\%} confidence intervals. Across the versions, we report average rank, Pearson $r$, and Kendall $\tau$ relative to the production ordering.

\paragraph{P4: Blinded human source discrimination.}
We sampled $N{=}\boldsymbol{100}$ \cdv{4} conversations, comprising \textbf{50} production and
\textbf{50} \snowglobe trajectories. \cdv{4} was selected because it was the most
recent version and the version most familiar to the annotators. \textbf{Seven}
domain experts contributed labels. After the same source-blind
normalization was applied to both groups, annotators saw dialogue text
without tool-call sequences, classified each item as production or
simulated, and reported confidence on a \textbf{1--5} scale.

\subsection{Open-weight model screening}
\label{sec:oss-method}

\paragraph{Models.} We test \textbf{29} configurations of the card management agent with over \textbf{16{,}000} simulated conversations, across model family, reasoning effort, and numerical format, including GPT-OSS-120B~\citep{openai2025gptoss120bgptoss20bmodel}, Nemotron-3-Super-120B-A12B~\citep{nvidia2026nemotron3superopen}, Qwen3.5-122B-A10B~\citep{qwen35122b}, and Qwen3.6-35B-A3B~\citep{qwen36_35b_a3b}.

\paragraph{Evaluation.} Building on the offline evaluation approach in Section \ref{subsec:card_delivery}, we use LLM-as-a-Judge to evaluate the following categories: (1) gathering proper inputs, (2) validity of card reissue, (3) information retrieval status, and (4) completeness of conversation, which are noted as ``Input'', ``Reissue'', ``Status'', and ``Complete'' in Table \ref{tab:oss-canonical}.

\section{Results}
\label{sec:results}

We characterize the simulator on CD versions, then report the offline CM analysis and the live A/B comparison, followed by a predeployment screen of open-weight
models.

\begin{figure}[!t]
  \centering
  \includegraphics[width=0.85\linewidth]{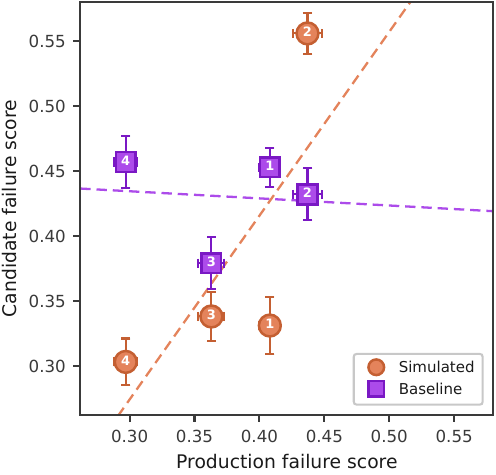}
  \caption{P3 -- \textbf{\snowglobe better preserves the production ordering than the baseline.} Aggregate E1--E5 failure scores for simulated and baseline conversations versus production (lower is better). Labels 1--4 identify CD versions. Error bars show 95\% conversation-clustered bootstrap confidence intervals.}
  \label{fig:p3-ranking}
\end{figure}

\begin{figure*}[t]
  \centering
      \includegraphics[width=1.0\linewidth]{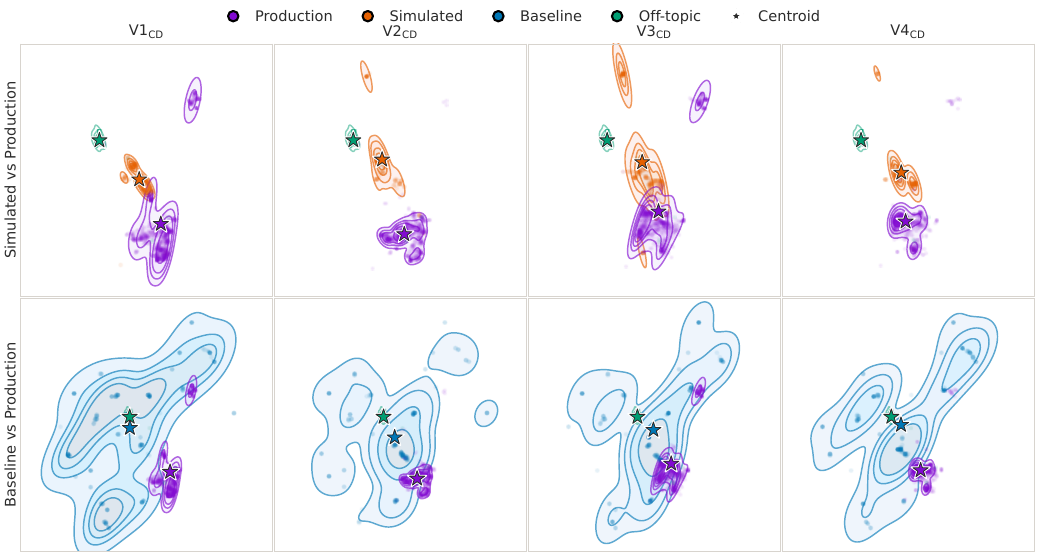}
\caption{P2 -- \textbf{\snowglobe samples are concentrated near production, while the baseline spans a
broader region extending toward the off-topic control.} Joint UMAP
projections compare \snowglobe with production (top) and the baseline
with production (bottom); the off-topic control is overlaid in both
rows. Haloed stars mark projected group means. }
  \label{fig:p2-scatter}
\end{figure*}

\subsection{Simulator characterization}
\label{sec:results-pillars}

\paragraph{P1: Conversation length.}
Appendix~\ref{app:p1-dists} reports both selected cumulative thresholds
for the \textbf{four} CD versions (Table~\ref{tab:p1-bins}) and the
complete distributions (Figure~\ref{fig:p1}). In the pooled samples, $\boldsymbol{22.4\%}$ of
simulated conversations contain at most \textbf{50} user-message words,
compared with $\boldsymbol{89.0\%}$ in production; $\boldsymbol{14.9\%}$ contain at least \textbf{200}
words, compared with $\boldsymbol{0.1\%}$ in production. Similarly, $\boldsymbol{16.4\%}$ of
simulated conversations have at most two user turns, compared with
$\boldsymbol{36.8\%}$ in production. The shares with at least six and eight turns
are $\boldsymbol{55.0\%}$ and $\boldsymbol{25.2\%}$ in simulation, versus $\boldsymbol{24.0\%}$ and $\boldsymbol{10.6\%}$
in production.

\paragraph{P2: Transcript-embedding proximity.}
For each CD version, the cosine distance between the
production and simulated transcript centroids is smaller than the
distance from either centroid to the off-topic control centroid
(Figure~\ref{fig:p2-centroid}). The same ordering holds for Euclidean
distance (Appendix~\ref{app:p2-euclidean},
Figure~\ref{fig:p2-centroid-euclidean}). Compared with the baseline,
\snowglobe is closer to production in two versions, tied in one at the
reported precision, and farther in one by cosine distance; by Euclidean
distance, it is closer in three versions and farther in one. Both
simulated groups remain closer to production than to the off-topic
control. Figure~\ref{fig:p2-scatter} provides the two-dimensional UMAP
projections.

\paragraph{P3: Evaluator-score association.}
Figure~\ref{fig:p3-ranking} shows that \snowglobe preserves the
production extremes, \cdv{4} best and \cdv{2} worst, although \cdv{1} and
\cdv{3} exchange positions. The baseline instead ranks \cdv{3} best and
\cdv{4} worst. \cdv{4} is best in $\boldsymbol{96.69\%}$ of \snowglobe resamples
and $\boldsymbol{0\%}$ of baseline resamples; \cdv{2} is worst in
$\boldsymbol{100\%}$ and $\boldsymbol{1.56\%}$, respectively.
Table~\ref{tab:p3-mean} in Appendix~\ref{app:p3-scores} reports the
aggregate scores and ranking-agreement metrics. The average-rank
statistic is \textbf{1.38} for \snowglobe and \textbf{1.62} for the baseline. The \snowglobe and production scores have Pearson
$\boldsymbol{r=0.74}$ and Kendall $\boldsymbol{\tau=0.67}$; their largest absolute difference is
for \cdv{2} (\textbf{0.556} versus \textbf{0.437}).

\paragraph{P4: Blinded human source discrimination.}
Annotators correctly identified \textbf{42} of \textbf{50} production conversations
($\boldsymbol{84.0\%}$) and \textbf{35} of \textbf{50} simulated conversations ($\boldsymbol{70.0\%}$). Thus, \textbf{15}
simulations were classified as production. Reported confidence was
approximately \textbf{3} on the \textbf{1--5} scale, including for correctly classified
items.

\subsection{Card Management: offline
and live results}
\label{sec:results-prod}

Applying the simulation recipe, \cmv{5} had the lowest unnecessary-transfer
failure rate ($\boldsymbol{16.0\%}$), compared with $\boldsymbol{22.4\%}$ for
\cmv{1}, $\boldsymbol{28.8\%}$ for \cmv{3}, and $\boldsymbol{34.0\%}$ for
\cmv{2} and \cmv{4}. This represents a \textbf{6.4}-percentage-point
reduction relative to \cmv{1}. In the subsequent A/B test, the CM arm had
SSR $\boldsymbol{4.90}$ percentage points above CD (\textbf{95\%} CI
$\boldsymbol{[4.08,\,5.71]}$) and tNPS $\boldsymbol{36.69}$ points above CD (\textbf{95\%} CI
$\boldsymbol{[32.81,\,40.57]}$)
(Table~\ref{tab:ab-results}). Following the test, the agent became
available to Nubank's customer base in Brazil.
A later test replaced the incumbent model in CM with the
open-weight candidate screened in Section~\ref{sec:results-oss},
Qwen3.5-122B-A10B with reasoning enabled. SSR rose by $\boldsymbol{+8.82}$
percentage points (\textbf{95\%} CI $\boldsymbol{[7.95,\,9.69]}$, $n=\boldsymbol{8.4\mathrm{K}}$) while
tNPS was statistically unchanged ($\boldsymbol{-1.21}$ points, \textbf{95\%} CI
$\boldsymbol{[-3.97,\,1.55]}$, $n=\boldsymbol{2.3\mathrm{K}}$),
and p95 latency fell by $\boldsymbol{25\%}$. The screened open-weight configuration
therefore matched the incumbent on satisfaction while improving
SSR and latency.

A batch of \textbf{100} simulated trajectories completed in under \textbf{10} minutes on
average; this measures generation runtime rather than end-to-end
iteration time. The \textbf{ten}-version CD development cycle,
including the \textbf{four} versions studied here, spanned \textbf{212} calendar days.
The \textbf{five} CM versions spanned \textbf{22} days. This corresponds to
$\boldsymbol{21.2}$ and $\boldsymbol{4.4}$ days per version, respectively, a ratio of $\boldsymbol{4.8}$.

\begin{table}[!t]
  \centering
  \caption{Live A/B outcomes for the \textbf{two} screened changes. Each panel
  reports differences against its own control. The p95 latency change is a
  point estimate from the serving stack.}
  \label{tab:ab-results}
  \small
  \begin{tabular}{@{}lrrr@{}}
    \toprule
    \textbf{Outcome} & \textbf{Difference} & \textbf{95\% CI} & $n$ \\
    \midrule
    \multicolumn{4}{@{}l}{\textit{Card Management vs.\ Card Delivery}} \\
    SSR (pp)      & $+4.90$  & $[4.08,\,5.71]$   & $27.8\mathrm{K}$ \\
    tNPS (pp)     & $+36.69$ & $[32.81,\,40.57]$ & $2.0\mathrm{K}$ \\
    \addlinespace
    \multicolumn{4}{@{}l}{\textit{Qwen3.5-122B-A10B (reasoning on) vs.\ incumbent}} \\
    SSR (pp)          & $+8.82$ & $[7.95,\,9.69]$  & $8.4\mathrm{K}$ \\
    tNPS (points) & $-1.21$ & $[-3.97,\,1.55]$ & $2.3\mathrm{K}$ \\
    p95 latency (\%)  & $-25$   & ---              & --- \\
    \bottomrule
  \end{tabular}
\end{table}

\subsection{Open-weight model screening}
\label{sec:results-oss}

Table \ref{tab:oss-canonical} reports \textbf{four} evaluator failure rates for the open-weight configurations. Performance by criterion, with no configuration uniformly outperforming the others. Qwen3.5 achieves the lowest input gathering (``Input'') failure rate with reasoning enabled (\textbf{0.4\%}) and the lowest conversation completeness (``Complete'') failure rate with reasoning disabled (\textbf{16.\%}). Qwen3.6 with reasoning enabled leads on card reissue validity (``Reissue'') (\textbf{14.0\%}), while GPT-OSS with medium reasoning has the lowest information retrieval status (``Status'') failure rate among open-weight configurations (\textbf{17.6\%}). The incumbent retains a substantial advantage on Status (\textbf{2.4\%}).

The effect of reasoning also varies by model and criterion. For Qwen3.6, enabling reasoning lowers all \textbf{four} reported mean failure rates. For Nemotron, moving from no to low reasoning reduces Complete failures from \textbf{60.8\%} to \textbf{18.0\%}, but increases Status failures from \textbf{21.2\%} to \textbf{36.4\%}. For Qwen3.5, reasoning reduces Input and Status failures while increasing Reissue and Complete failures. These results imply model-specific benefits from additional reasoning.

We selected Qwen3.5-122B-A10B with reasoning enabled as a starting point for prompt optimization based on its low Input failure rate and competitive Reissue and Complete results. Relative to the incumbent, its
Input and Complete failure rates are lower by \textbf{3.2} and \textbf{15.0} percentage points, respectively, while Reissue is similar (\textbf{25.4\%} versus \textbf{25.2\%}). The higher Status failure rate
(\textbf{35.2\%} versus \textbf{2.4\%}) identified a weakness for further
refinement. Subsequent prompt optimization reduced
failures on this criterion. The offline results therefore supported Qwen3.5-122B-A10B as a candidate for live evaluation, without establishing it as the strongest configuration across all criteria. Table~\ref{tab:ab-results} reports the subsequent live A/B outcomes.


\begin{table}[t]
\centering
\caption{Canonical evaluation failure rates (\%) on simulated card management
conversations ($\downarrow$). Bold and underlined values indicate
the lowest and second-lowest means among open-weight configurations,
respectively.}
\label{tab:oss-canonical}

\small
\setlength{\tabcolsep}{2pt}
\renewcommand{\arraystretch}{1.08}

\resizebox{\columnwidth}{!}{%
\begin{tabular}{@{}ll*{4}{c}@{}}
\toprule
\textbf{Configuration} & \textbf{Reasoning} &
\multicolumn{4}{c}{\textbf{Canonical Evaluations} ($\downarrow$)} \\
\cmidrule(lr){3-6}
& & \textbf{Input} & \textbf{Reissue}
& \textbf{Status} & \textbf{Complete} \\
\midrule
GPT-5.2 (Incumbent) & \textemdash
& $3.6_{3.0}$ & $25.2_{7.9}$ & $2.4_{3.3}$ & $37.6_{10.1}$ \\
\midrule
\multirow{2}{*}{Qwen3.6-35B-A3B} & Off
& $1.6_{1.7}$ & $24.0_{3.7}$ & $29.2_{9.0}$ & $25.2_{4.1}$ \\
& On
& $1.2_{1.8}$ & $\mathbf{14.0}_{3.7}$
& $25.6_{3.3}$ & $\underline{17.6}_{7.4}$ \\
\addlinespace[2pt]
\multirow{2}{*}{Qwen3.5-122B-A10B} & Off
& $\underline{0.8}_{1.1}$ & $24.8_{7.7}$
& $38.8_{6.9}$ & $\mathbf{16.0}_{3.7}$ \\
& On
& $\mathbf{0.4}_{0.9}$ & $25.4_{7.7}$
& $35.2_{10.1}$ & $22.6_{12.2}$ \\
\addlinespace[2pt]
\multirow{3}{*}{\shortstack[l]{Nemotron-3-Super-\\120B-A12B}} & Off
& $6.4_{2.2}$ & $32.0_{7.6}$ & $21.2_{5.0}$ & $60.8_{12.5}$ \\
& Low
& $3.2_{2.3}$ & $20.0_{6.3}$ & $36.4_{2.6}$ & $18.0_{4.2}$ \\
& High
& $1.2_{2.7}$ & $24.0_{4.9}$ & $30.0_{6.3}$ & $19.2_{5.4}$ \\
\addlinespace[2pt]
\multirow{3}{*}{GPT-OSS-120B} & Low
& $2.0_{2.4}$ & $\underline{18.4}_{3.6}$
& $28.4_{6.2}$ & $\underline{17.6}_{4.3}$ \\
& Medium
& $3.2_{2.3}$ & $26.0_{7.7}$
& $\mathbf{17.6}_{7.9}$ & $20.8_{4.6}$ \\
& High
& $2.4_{0.9}$ & $28.0_{7.7}$
& $\underline{19.6}_{3.0}$ & $18.8_{6.1}$ \\
\bottomrule
\end{tabular}%
}
\end{table}

\subsection{Deployment lessons and limitations}
\label{sec:lessons-limitations}

Simulation supported safer, faster iteration: batches of \textbf{100} trajectories
ran in under \textbf{ten} minutes, evaluators surfaced behavioral regressions,
and the screen caught tool-call failures and a serving configuration
without a tool-call parser. Most engineering effort went into
integration: the staging agent reused live MCP schemas while stateful tools returned schema-compatible synthetic responses. The agent retained tool control and
sequencing, but schemas, examples, authentication, and simulator profiles
required ongoing alignment.

The approach deliberately stops at the tool boundary. Stateful or
side-effecting tools are mocked, while read-only dependencies such as
knowledge-base retrieval may remain live; backend behavior, latency,
persistent state, and side effects thus remain untested. Simulation
traces must also match production schemas, identifiers, and telemetry, or
export and reconciliation work can erase the iteration gains.

\section{Conclusion}
\label{sec:conclusion}

In this paper, we present a hypothesis-driven approach to using simulation as a pre-deployment screening layer for CX agents. At Nubank, simulation has become an integral part of our development life cycle, enabling teams to explore changes, identify behavioral failures, and refine candidates before customer exposure. Its value lies in supporting development decisions, not in perfectly reproducing production. By combining simulation with live evaluation, this workflow supports broader experimentation while keeping deployment decisions grounded in observed customer outcomes.

\bibliography{references}
\bibliographystyle{icml2026}

\clearpage
\appendix

\section{How \snowglobe infers simulation properties}
\label{app:inference}

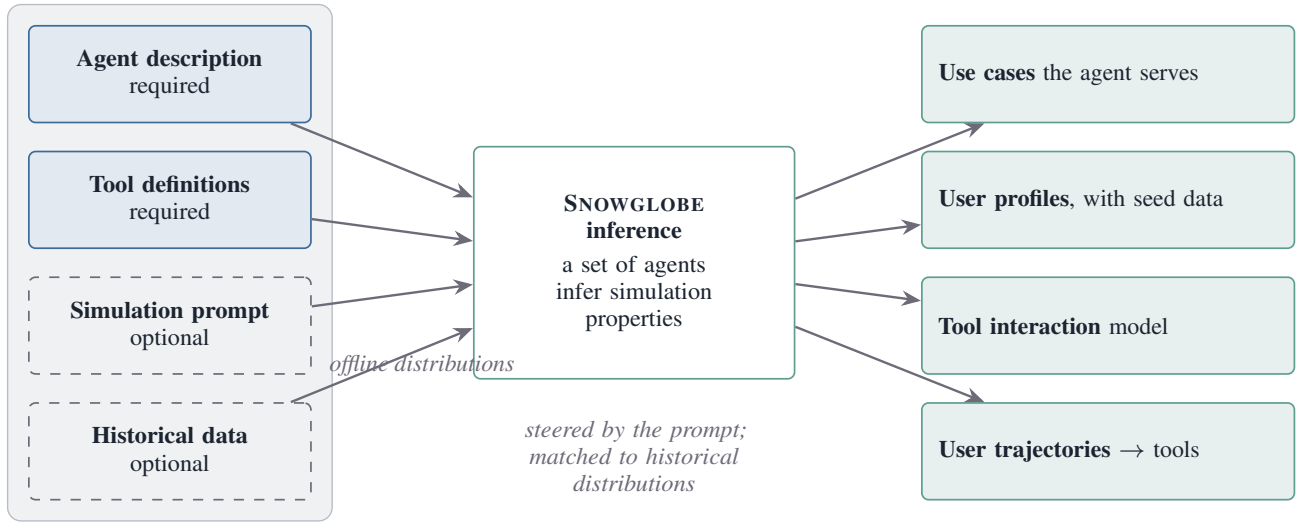
\begin{figure*}[!htb]
  \centering
  \resizebox{\linewidth}{!}{
\begin{tikzpicture}[node distance=8mm]
  \tikzset{
    inp/.style ={intbox, text width=2.6cm, align=center, minimum height=1.0cm},
    optinp/.style={inp, dashed, draw=edge, text=ink, fill=clbg},
    infer/.style={box, fill=white, text width=3.0cm, align=center,
                  minimum height=2.4cm},
    prop/.style ={hlbox, text width=3.5cm, align=left, minimum height=1.0cm,
                  inner sep=5pt},
    note/.style ={font=\scriptsize\itshape, text=edge, align=center},
  }
  \node[inp]    (AD) at (0,3.9) {\textbf{Agent description}\\{\scriptsize required}};
  \node[inp]    (TD) at (0,2.6) {\textbf{Tool definitions}\\{\scriptsize required}};
  \node[optinp] (SP) at (0,1.3) {\textbf{Simulation prompt}\\{\scriptsize optional}};
  \node[optinp] (HD) at (0,0.0) {\textbf{Historical data}\\{\scriptsize optional}};
  \node[infer] (INF) at (4.8,1.95)
    {\textbf{\textsc{Snowglobe}}\\\textbf{inference}\\[2pt]
     {\scriptsize a set of agents\\infer simulation\\properties}};
  \node[prop] (UC) at (9.7,3.9) {\textbf{Use cases} the agent serves};
  \node[prop] (UP) at (9.7,2.6) {\textbf{User profiles}, with seed data};
  \node[prop] (TI) at (9.7,1.3) {\textbf{Tool interaction} model};
  \node[prop] (TR) at (9.7,0.0) {\textbf{User trajectories} $\rightarrow$ tools};
  \foreach \n in {AD,TD,SP} \draw[arr] (\n) -- (INF);
  \draw[arr] (HD) -- node[lbl, below, pos=.72]{offline distributions} (INF);
  \foreach \n in {UC,UP,TI,TR} \draw[arr] (INF) -- (\n);
  \node[note, below=3mm of INF, text width=3.4cm]
    {steered by the prompt;\\matched to historical\\distributions};
  \begin{scope}[on background layer]
    \node[cl, fit=(AD)(TD)(SP)(HD)] (inb) {};
  \end{scope}
  \node[cltitle, above=1pt of inb.north west, anchor=south west] {inputs};
\end{tikzpicture}}
  \caption{Property inference in \snowglobe. From a required agent
  description and tool definitions, optionally augmented with a
  simulation prompt and historical data, a set of agents infers the
  use cases, user profiles and their seed data, the tool interaction
  model, and the user trajectories over the tools. Historical data is
  processed offline into features that can condition later runs.}
  \label{fig:inference}
\end{figure*}

Before it can interact with an agent in a meaningful way, \snowglobe
needs an approximation of that agent's objectives, its users, and its
behavior. The minimal input is a description of the agent together
with the definitions of its tools. From these, a set of agents infers
the properties a simulation depends on (Figure~\ref{fig:inference}):
the use cases the agent serves, profiles of synthetic users
along with
the data those users would carry, how the tools relate to one another
and to the surrounding system, and the user trajectories that would
call on each tool. Some of these properties are resolved once during agent
onboarding and others are resolved when a simulation begins.

\textbf{Two} optional inputs reorient these properties. A simulation prompt
narrows a run to a chosen set of use cases or behaviors while leaving
the rest of the inferred picture intact. Real historical
conversations, when supplied, are processed offline to extract use
cases, linguistic styles, and their distributions, which are stored and
used to condition later runs.

\section{Tool boundary mocking}
\label{app:mocking}

Exercising the agent's tools inside a simulation raises a problem of
its own. The simulated environment must not access production
data through stateful tools, yet it has to return data through those
tools so the
agent can run end to end, and that data must stay consistent with the
state of the conversation and with the results of earlier tool calls. Testing
the tools themselves is a separate concern, better served by other
methods, and is not an aim here.

\begin{figure*}[!htb]
  \centering
  \resizebox{\linewidth}{!}{
\definecolor{stop}{HTML}{C0392B}
\begin{tikzpicture}[node distance=10mm]
  \tikzset{
    party/.style ={box, text width=2.6cm, align=center, minimum height=1.4cm},
    blackbox/.style={rectangle, rounded corners=3pt, draw=ink, fill=ink,
                     text=white, align=center, font=\scriptsize,
                     text width=3.0cm, minimum height=1.6cm, inner sep=6pt},
    feed/.style  ={box, text width=2.7cm, align=center, minimum height=1.2cm},
    gone/.style  ={rectangle, rounded corners=3pt, draw=edge, dashed,
                   fill=clbg, text=edge, align=center, font=\scriptsize,
                   text width=4.6cm, minimum height=1.3cm, inner sep=6pt},
    note/.style  ={font=\scriptsize\itshape, text=edge, align=center},
  }
  \node[party, simbox] (USER)  {\textbf{Simulated user}\\{\scriptsize persona-driven}};
  \node[party, intbox, right=10mm of USER] (AGENT)
        {\textbf{Agent under test}\\{\scriptsize user-provided}};
  \draw[{Stealth}-{Stealth}, draw=edge, line width=.7pt]
        (USER)--node[note, above]{turns}(AGENT);

  \node[blackbox, right=34mm of AGENT] (MOCK)
        {\textbf{Mock tool layer}\\[1pt]{\scriptsize answers one tool\\call at a time}};
  \draw[arr, draw=ink] ([yshift=4pt]AGENT.east) --
        node[note, above]{tool call} ([yshift=4pt]MOCK.west);
  \draw[arr, draw=ink] ([yshift=-4pt]MOCK.west) --
        node[note, below]{mocked result} ([yshift=-4pt]AGENT.east);

  \node[feed, simbox, above=13mm of MOCK, xshift=-16mm] (PERS)
        {\textbf{Persona}\\{\scriptsize from the sim ---\\the world it implies}};
  \node[feed, planbox, above=13mm of MOCK, xshift=16mm] (PROF)
        {\textbf{Agent profile}\\{\scriptsize rules each tool\\must follow}};
  \draw[arr] (PERS) -- (MOCK.north -| PERS);
  \draw[arr] (PROF) -- (MOCK.north -| PROF);

  \node[gone, below=16mm of MOCK] (BK)
        {Real backend $\cdot$ databases\\$\cdot$ services $\cdot$ persistent state};
  \draw[dashed, edge] (MOCK.south) --
        node[pos=.52, fill=white, inner sep=1pt, text=stop, font=\normalsize\bfseries]{$\times$}
        node[note, right, pos=.52, xshift=3pt]{never invoked} (BK.north);
\end{tikzpicture}}
  \caption{Tool-boundary mocking. As the simulated user and the agent
  under test exchange turns, each tool call the agent makes is answered
  by a mock tool layer parameterised by the persona and the agent
  profile. Only the tool boundary is mocked.}
  \label{fig:mocking}
\end{figure*}
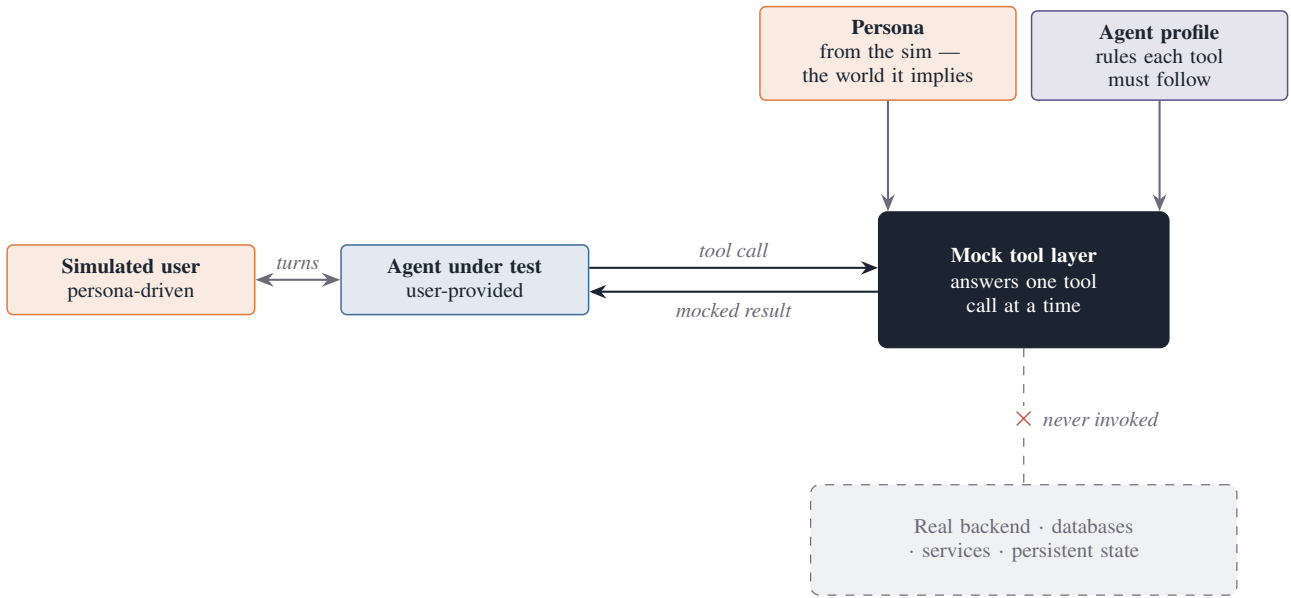

\snowglobe meets these constraints without ever maintaining a database
of state (Figure~\ref{fig:mocking}). During onboarding, it absorbs the tool
definitions and builds a functional model of how the tools relate and
how data must be shaped to satisfy each request. This model is called an 
agent profile. Using an agent profile, an orchestrator produces persona
seeds. These are partial seed states paired with a trajectory for how that data
should evolve over a conversation. Pesrona seeds are projected onto personas at
runtime to ground their objectives.

At simulation time, a call to one of the agent's tools is routed to the
persona agent handling that conversation rather than to the real
implementation. Using its partial state and the consistency rules in
the agent profile, the persona agent returns a response that fits the
tool's output shape, and that response is passed back to the agent
under test. Because no shared world state is instantiated, each
conversation runs in its own sandbox. \snowglobe refers to this as
tool-boundary mocking: only the individual tool calls at the boundary
are answered, never an entire backend or database.

\section{Conversation length distributions}
\label{app:p1-dists}
\begin{figure*}[!htb]
  \centering
    \includegraphics[width=1.0\linewidth]{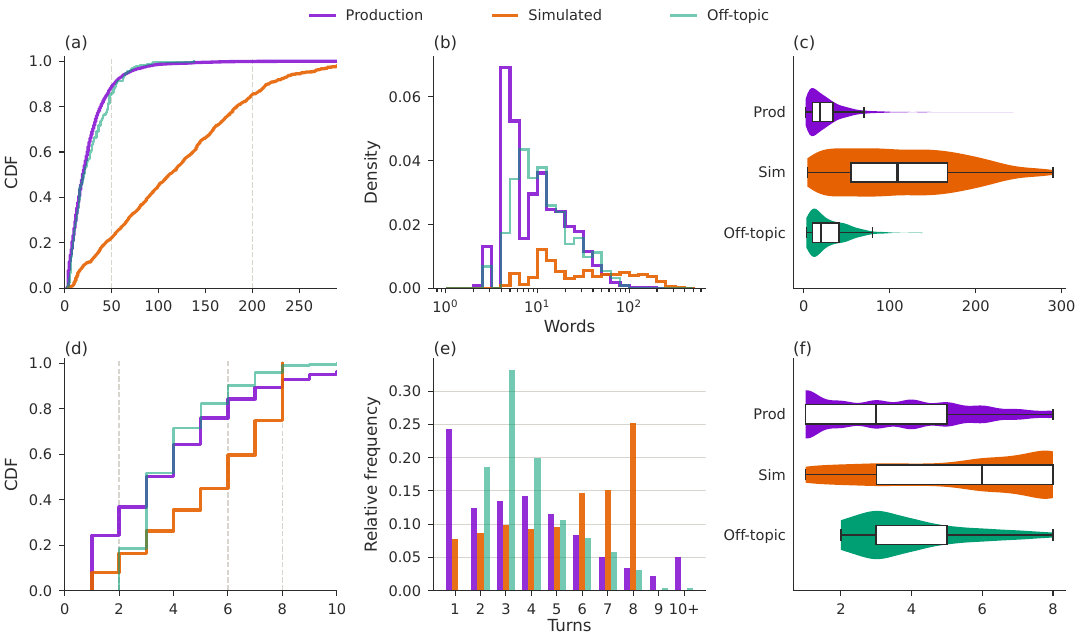}
    \caption{P1 -- \textbf{\snowglobe simulated conversations are substantially longer than production and off-topic conversations in both user-message words and user turns.} The top row shows word count and the bottom row user turns: ECDFs (a,d), density or frequency (b,e), and violins with nested boxplots (c,f). Panel~(e) pools values $\geq \boldsymbol{10}$ into the \texttt{\textbf{10+}} bin.}

  \label{fig:p1}
\end{figure*}

Figure~\ref{fig:p1} shows the full P1 distributions of
user-message words and user turns, complementing the thresholds in
Table~\ref{tab:p1-bins}. Length is user-side only. Groups are
production traffic pooled over \cdv{1}--\cdv{4} ($n=\boldsymbol{4000}$),
\textsc{Snowglobe} simulations of the same variants ($n=\boldsymbol{1000}$), and an
off-topic control ($n=\boldsymbol{226}$). The retained \textbf{226}-item control is encoded in the plotted
artefact and historical experiment code, but the filtering from a
documented upstream \textbf{1,000}-item random control sample is not recorded. Table \ref{tab:p1-bins} shows summarized results in bins for easy comparison of extreme values.

Production user text is short (median \textbf{19} words; mean \textbf{25.8}). The
control matches that scale (median \textbf{20}; mean \textbf{27.5}); simulated users do
not (median \textbf{111}; mean \textbf{119}). Control shares at the Table~\ref{tab:p1-bins}
word cuts are \textbf{85.8\%} ($\leq \boldsymbol{50}$) and \textbf{0\%} ($\geq \boldsymbol{200}$).
Because the control is production traffic rather than
LLM-generated user text, it cannot determine whether the heavy
simulated tail is specific to this configuration or common to LLM user
simulators.

In production, $\boldsymbol{24.3\%}$ of conversations have a single user
turn, and the median is \textbf{three} turns. Control never has a \textbf{one}-turn chat
and peaks at \textbf{three}
turns (\textbf{33.2\%}). Simulated conversations last longer (median \textbf{6}) and
hit a hard cap: \textbf{25.2\%} have exactly \textbf{eight} user turns, and none have
more---which is why every simulated conversation has fewer than
\textbf{10} user turns, the \textbf{95}th-percentile value in production, despite the
higher simulated median.
Control shares at the table cuts are \textbf{18.6\%} ($\leq \boldsymbol{2}$), \textbf{17.7\%}
($\geq \boldsymbol{6}$), and \textbf{4.0\%} ($\geq \boldsymbol{8}$).

\begin{table*}[!htb]
  \centering
  \caption{P1 -- Selected cumulative
  conversation-length thresholds for \cdv{1}--\cdv{4}.
  Values are percentages of conversations meeting each criterion; rows
  may overlap. Word counts include user messages only.  }
  \label{tab:p1-bins}
  \small
  \begin{tabular}{lcc@{\hspace{1.2em}}lcc}
    \toprule
    \multicolumn{3}{c}{User-message words per conversation} &
    \multicolumn{3}{c}{User turns per conversation} \\
    \cmidrule(lr){1-3} \cmidrule(lr){4-6}
    Criterion & Production &
    Simulated & Criterion &
    Production & Simulated \\
    \midrule
    $\leq 50$        & 89.0 & 22.4 & $\leq 2$ & 36.8 & 16.4 \\
    $\geq 200$ words &  0.1 & 14.9 & $\geq$~6 & 24.0 & 55.0 \\
                     &      &      & $\geq$~8 & 10.6 & 25.2 \\
    \bottomrule
  \end{tabular}
\end{table*}

\section{Euclidean transcript-centroid distances}
\label{app:p2-euclidean}

\begin{figure*}[!htb]
  \centering
  \includegraphics[width=0.95\linewidth]{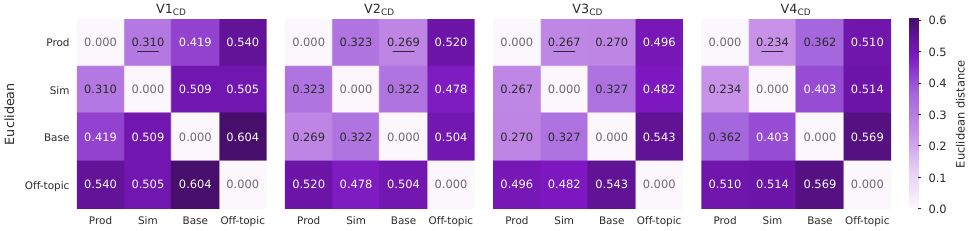}
  \caption{P2 -- \textbf{In all versions, both simulated groups are
closer to production than the off-topic control; \snowglobe is no
farther from production than the baseline in three versions.}
Euclidean distances between whole-conversation embedding centroids.
Underlined values mark the nearest non-production centroid.}
  \label{fig:p2-centroid-euclidean}
\end{figure*}

Figure~\ref{fig:p2-centroid-euclidean} reports the Euclidean-distance
counterpart to the cosine-distance result in
Figure~\ref{fig:p2-centroid}.

\section{Evaluator-score association details}
\label{app:p3-scores}

Table~\ref{tab:p3-mean} reports the P3 aggregate scores and
ranking-association statistics. Each conversation contributes \textbf{five}
binary failure outcomes, \textbf{one} per evaluator. Within each source--version
pool, we first average each evaluator across conversations and then
average the \textbf{five} evaluator means; lower aggregate scores indicate fewer
failures.

The \textbf{95\%} intervals use $\boldsymbol{10{,}000}$ conversation-clustered bootstrap
replicates. Each replicate resamples conversation rows with replacement,
preserving the \textbf{five} outcomes for each sampled conversation, and
recomputes the aggregate score. Production defines the reference
ordering. Average rank summarizes closeness to that order (lower is
better), Pearson $r$ measures linear association between version-level
scores, and Kendall $\tau$ measures rank agreement.

The extreme-rank analysis uses the same resampling procedure. Each replicate ranks the \textbf{four} versions by
aggregate score, with ties sharing credit equally. The reported
frequencies estimate only how often \cdv{4} is best and \cdv{2} is worst.

\begin{table*}[!htb]
  \centering
  \caption{P3 -- Aggregate E1--E5 failure scores by version (lower is
  better), reported as means $\pm$ \textbf{95\%} conversation-clustered
  bootstrap half-widths. Production defines the reference ranking;
  average rank, Pearson $r$, and Kendall $\tau$ measure agreement with
  it.}
  \label{tab:p3-mean}
  \small
  \begin{tabular*}{\textwidth}{@{\extracolsep{\fill}}l cccc ccc@{}}
    \toprule
    \textbf{Source} & \textbf{\boldmath\cdv{1}} & \textbf{\boldmath\cdv{2}} & \textbf{\boldmath\cdv{3}} & \textbf{\boldmath\cdv{4}} &
    \textbf{Avg.\ rank} & \textbf{Pearson} $r$ & \textbf{Kendall $\tau$} \\
    \midrule
    Baseline & $0.453_{\pm 0.015}$ & $0.432_{\pm 0.020}$ & $0.379_{\pm 0.020}$ & $0.457_{\pm 0.020}$ & $1.62$ & $-0.09$ & $-0.33$ \\
    Simulated & $0.331_{\pm 0.022}$ & $0.556_{\pm 0.016}$ & $0.338_{\pm 0.019}$ & $0.303_{\pm 0.018}$ & $\textbf{1.38}$ & $\textbf{0.74}$ & $\textbf{0.67}$ \\
    Production & $0.408_{\pm 0.008}$ & $0.437_{\pm 0.011}$ & $0.363_{\pm 0.010}$ & $0.297_{\pm 0.009}$ & $1.00$ & $1.00$ & $1.00$ \\
    \bottomrule
  \end{tabular*}
\end{table*}

\section{Open-weight screening: reasoning and quantization}
\label{app:oss-screening}

Because a simulated arm costs no customer exposure, the screen can
afford to sweep configuration settings that would otherwise be argued
from intuition. This appendix reports the sweep summarised in
Section~\ref{sec:results-oss}, run on Nemotron-3-Super-120B-A12B across
\textbf{three} quantizations and \textbf{three} reasoning settings. The figure omits the
incumbent: the composite is a mean over evaluators built for this
agent, and is used to compare a model against itself under different
settings rather than to rank candidates against production.

\begin{figure*}[!htb]
\centering
\includegraphics[width=0.56\linewidth]{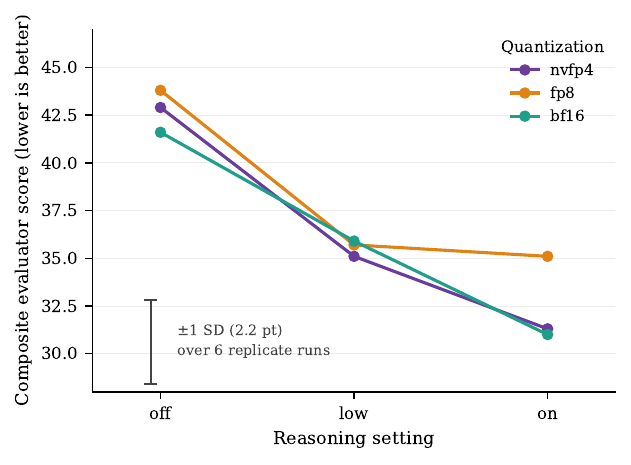}
\caption{Reasoning and quantization sweep for
Nemotron-3-Super-120B-A12B: composite evaluator score (lower is better)
against reasoning setting, \textbf{one} series per quantization. The bar is
$\pm \boldsymbol{1}$ standard deviation of the composite measured over \textbf{six} replicate
runs of a fixed configuration, and is a scale reference rather than a
comparison against any arm.}
\label{fig:oss-quant}
\end{figure*}

Figure~\ref{fig:oss-quant} shows the result. Reasoning effort moves the
composite monotonically and by a large margin, from $\boldsymbol{42.9}$ with
reasoning off to $\boldsymbol{31.3}$ with it on at \texttt{nvfp4}, and the mechanism
is retrieval: the rate at which the model fetches the company record
the task requires rises from $\boldsymbol{16\%}$ to $\boldsymbol{94\%}$ over the same range.
Quantization does not move it. Across \texttt{nvfp4}, \texttt{fp8}, and
\texttt{bf16} the spread is $\boldsymbol{2.2}$, $\boldsymbol{0.8}$, and $\boldsymbol{4.1}$ points at reasoning
off, low, and on, against a $\boldsymbol{2.2}$-point standard deviation measured
over \textbf{six} replicate runs of a fixed configuration. At this sample size
the \textbf{three} quantizations are indistinguishable, while the reasoning
setting moves the composite by roughly $\boldsymbol{11}$ points.

\section{Baseline simulator prompts}
\label{app:baseline-prompts}

Both simulator models (user and tools) are \texttt{gpt-5.6-sol} with
\texttt{reasoning\_effort=high}.
Brace placeholders mark run-specific context concatenated at inference
time: agent description, tool JSON schemas, product-requirements text,
and tool-call examples.

\subsection{Persona model}

The persona is given \textbf{one} function tool, \texttt{end\_conversation},
so it can trigger the end of the simulation.
The first user message is shown below.
Later turns reuse the same system prompt and append the dialogue
(user and assistant text only).

\begin{promptcard}{Persona system prompt}
You are a Brazilian Nubank customer chatting
in the Nubank app.

Stay in character. Write in informal Brazilian
Portuguese, like a real customer on a phone
keyboard: short messages, occasional typos,
no markdown.

You are talking to a support chatbot. You do
not work at Nubank. You do not know internal
tool names, IDs, or policies beyond what is
listed below.

Your persona:
{persona_card}

Rules:
- Send one customer message at a time.
- If the bot resolved your issue, or you are
  done, call end_conversation.
- If the bot transfers you to a human, call
  end_conversation.
- Do not invent that a human already joined.
- Maximum 10 of your messages in the whole
  conversation.

## Agent description
{description}

## Tools the agent can use (description + schema)
{tools_schema}

## Product requirements (meta-knowledge)
{prd}

## Tool examples
{tool_examples}
\end{promptcard}

\begin{promptcard}{Persona user message}
Write your first message to the Nubank chatbot now.
\end{promptcard}

Customer cards are sampled from a seeded generator.

\begin{promptnotecard}{Persona card}
Nome: \{name\}. Cidade: \{city\}. Problema: \{issue\}.\\
Tom: \{tone\}. Voc\^e \'e cliente Nubank no Brasil e\\
est\'a no chat do app.
\end{promptnotecard}

\subsection{Tool-result model}

The mock backend is a second completion (no tools).
The user message is the pending tool call.

\begin{promptcard}{Tool-result system prompt}
You are an LLM tool-result simulator for a
Nubank card-delivery conversation.
Generate a realistic JSON result for the
given tool call.
Return ONLY the tool output (JSON if
possible). No preamble.
Stay consistent with earlier results in this
conversation when the same IDs appear.

## Agent description
{description}

## Tool description + schema
{tools_schema}

## Product requirements (meta-knowledge)
{prd}

## Tool examples
{tool_examples}
\end{promptcard}

\begin{promptcard}{Tool-result user message}
Generate a result for this tool call.
Tool: {tool_name}
Arguments: {arguments}
\end{promptcard}

\makeatletter\setlength{\@fptop}{0pt}\setlength{\@dblfptop}{0pt}\makeatother\end{document}